\pdfoutput=1
\documentclass[11pt]{article}
\usepackage[utf8]{inputenc}
\usepackage[T1]{fontenc}
\usepackage{mathptmx}
\usepackage[margin=1in]{geometry}
\usepackage{microtype}
\usepackage{amsmath}
\usepackage{graphicx}
\usepackage{booktabs}
\usepackage{tabularx}
\usepackage{array}
\usepackage{xcolor}
\usepackage[round,authoryear]{natbib}
\usepackage[hidelinks,pdftitle={Information Access and Gain-Loss Framing in Six Large Language Models and Human Decision Makers},pdfauthor={Aadithya Chandramouli and Jeremy M. Gernand}]{hyperref}
\usepackage{caption}
\title{Information Access and Gain-Loss Framing in Six Large Language Models and Human Decision Makers}
\author{
  Aadithya Chandramouli\\
  School of Electrical Engineering and Computer Science\\
  College of Engineering, The Pennsylvania State University
  \and
  Jeremy M. Gernand\\
  College of Earth and Mineral Sciences\\
  The Pennsylvania State University
}
\date{}
\begin{document}
\maketitle
\begin{abstract}
A risky choice can be described by what will be saved or what will be lost, and people often choose differently depending on that wording. We test whether six large language models (LLMs) show this gain-loss framing effect, and whether it shrinks when they can first read more about the options. GPT-4.1, GPT-4o, GPT-4o-mini, Gemini 2.5 Flash, Qwen3 235B and Llama 4 Maverick made 11,519 choices across four risk scenarios, each in its original wording, two paraphrases and a mirrored version in the opposite wording. Before choosing, a model could open up to 0, 2, 3 or 4 of seven information cards. With no cards, all six models showed large framing effects: the share choosing the safe option was 27 to 79 percentage points lower under loss than under gain wording. The gap was smaller at 2 cards than with no cards in every model. Against a 10-point threshold set in our analysis plan, GPT-4.1 fell within it at 4 cards and Qwen3 only at 2 cards; in the plan's pooled comparison over all versions, GPT-4.1 was within it from 2 cards and GPT-4o from 3. GPT-4o-mini and Llama 4 Maverick kept gaps of 28 to 66 points with cards, and for GPT-4o-mini and Qwen3 the gap grew again from 2 to 4 cards. Part of the drop from 0 to 2 cards may reflect a change in prompt format, and most card content counted against the risky option, which may favor the safe choice. In an exploratory comparison with 198 students who saw the same scenarios and cards but opened cards one at a time and chose how many, decisions made after opening 0 or 1 cards showed a 19-point difference between the gain scenarios, asked first, and the loss scenarios, asked later, and no clear difference appeared in decisions made after opening more, though this design cannot show that cards reduced framing in people. With these cards, in the balanced comparison 17 of 18 card gaps were below the model's no-card gap, but few were within the threshold.
\end{abstract}

\section{Introduction}

The same risky choice can be described in two ways. A response program can save 200 of 600 people for certain, or it can let 400 of 600 people die for certain. The outcomes are identical, yet people tend to prefer the sure (safe) option when outcomes are described as gains and to gamble when they are described as losses \citep{tversky1981framing}. This gain-loss framing effect has been replicated many times \citep{kuhberger1998influence}, and biases of this kind matter wherever risks are assessed and communicated \citep{montibeller2015cognitive,gernand2023set}.

As LLMs are used to answer questions and support decisions, it matters whether their choices also depend on wording. \citet{binz2023using} found that GPT-3 showed a framing effect on classic vignettes, and \citet{echterhoff2024cognitive} found a range of cognitive biases, including framing, in LLM decision making. Two recent studies test gain-loss framing in current models: one in risky choices, comparing reasoning and conversational models with a matched human experiment \citep{ge2026mind}, and one in high-stakes advice, where the same probability is stated as a gain or a loss \citep{zhang2026same}. Most such tests give a model a short vignette and ask for a choice. In practice, a person or a model can often read more about each option before deciding.

We ask whether access to option-specific information changes how much the wording matters, and how much access is needed before the difference falls below a practical threshold. A decision maker who reads more about each option might rely less on how the outcomes are worded.

We study this with an information-card task. Each scenario offers a safe and a risky option, worded as either a gain or a loss. Before choosing, the decision maker can open up to a set number of the seven cards that describe the two options. We ran six LLMs at card limits of 0, 2, 3 and 4, in the original wording, two paraphrases and a mirrored version of each scenario, and compared them with 198 students who saw the original scenarios and the same cards.

We make three contributions:

\begin{enumerate}
  \item We introduce an information-card version of the gain-loss task, in which a model chooses which of seven cards to read, and measure the framing gap in six LLMs at four card limits against a 10-point threshold set in a written analysis plan.
  \item We find that gaps were smaller with card access in most models but rarely within the threshold, and that in two models the gap was larger at higher card limits.
  \item We use mirrored scenarios to compare the two frames within each topic and paraphrases to measure how much each model's results depend on the exact sentences.
\end{enumerate}

We also report an exploratory comparison with people who saw the same scenarios and cards.

\section{Related work}

\paragraph{Framing in human risky choice.} In prospect theory, people evaluate outcomes as gains or losses from a reference point and tend to avoid risk among gains and seek risk among losses \citep{kahneman1979prospect}. \citet{tversky1981framing} showed that describing the same outcomes in gain or loss terms reverses many people's preferences. \citet{kuhberger1998influence} pooled framing studies in a meta-analysis and found a reliable effect whose size depends on how the task is designed. \citet{levin1998all} separated risky-choice framing from attribute framing and goal framing; our scenarios are risky-choice framing. \citet{schultemecklenbeck2014out} tracked which information people opened in the Asian disease problem and found that search did not differ between the gain and loss frames, even where the frame changed choices. When the chance of success is given as a range instead of an exact number, people face ambiguity as well as risk \citep{ellsberg1961risk}. Two of our scenarios give an exact chance (1 in 3) and two give a range (10\% to 50\%). In risk analysis, cognitive biases affect how hazards are judged \citep{montibeller2015cognitive}, and engineering students show measurable biases in risk estimation and decision preferences \citep{gernand2023set}.

\paragraph{Cognitive biases in LLMs.} \citet{binz2023using} applied tasks from cognitive psychology to GPT-3, including vignettes on framing and tasks that involve searching for information, and found human-like behavior on several of them. \citet{hagendorff2023human} found that intuitive reasoning errors on semantic illusions and cognitive-reflection problems appeared as models grew larger and mostly disappeared in later chat models. \citet{suri2024large} reported human-like decision heuristics in GPT-3.5; its framing task concerned how positively or negatively an attribute is described, a different construct from the risky-choice framing studied here. \citet{macmillanscott2024irrationality} found that model errors on such tasks are highly inconsistent and mostly unlike human biases. \citet{echterhoff2024cognitive} tested several biases, including framing, in LLM decision making and evaluated ways to reduce them. Two recent studies test gain-loss framing in current models. \citet{ge2026mind} compared 20 models in risky choices and found that, in the configurations they tested, conversational models were sensitive to the frame while reasoning models were not; conversational models also chose differently when the same prospects were described or shown as past outcomes. \citet{zhang2026same} studied advice rather than a choice between a sure option and a gamble: the same probability was stated as a gain or a loss, and across seven models this changed the recommendation category in about a third of matched sets of high-stakes prompts, mostly through shifts to or from an uncertain answer rather than direct reversals. A step they call probability canonicalization reduced this sensitivity. Our study adds a narrower manipulation: varying how many self-selected, option-specific cards a model may read, in a framing task with mirrored scenarios.

\paragraph{Prompt wording and training data.} LLM performance can change substantially with small changes to prompt formatting that do not change meaning \citep{sclar2024quantifying}, so a result obtained with one prompt may not hold for another. We use paraphrases to check sensitivity to wording, which is related to, though not the same as, the formatting changes studied by Sclar and colleagues. Well-known problems such as the ``Asian disease'' framing task may also appear in training data, a general concern for LLM evaluation \citep{sainz2023nlp}. We cannot test whether any model saw our exact scenarios.

\section{Methods}

\subsection{Scenarios}

Each of the four scenarios describes 600 units at risk and offers a safe option, which keeps one third of the units for certain, and a risky option, which either keeps everything or loses everything (Table 1). Two scenarios are worded as gains and two as losses. Two give an exact chance of success and two give a range. The full text is in the Supplementary information.

\begin{table}[ht]
\centering
\footnotesize
\caption{The four original scenarios.}
\label{tab:1}
\setlength{\tabcolsep}{4pt}
\begin{tabularx}{\linewidth}{>{\raggedright\arraybackslash}X l >{\raggedright\arraybackslash}X>{\raggedright\arraybackslash}X}
\toprule
Scenario & Wording & Safe option & Risky option \\
\midrule
Q1 Disease outbreak (600 people) & Gain & 200 people are saved for certain & 1-in-3 chance all 600 are saved \\
Q2 Reserve fund (\$600,000) & Gain & \$200,000 is preserved for certain & 10\% to 50\% chance all is preserved \\
Q3 Storm and municipal equipment (600 units) & Loss & 400 units are lost for certain & 1-in-3 chance nothing is lost \\
Q4 Wildfire (600 hectares) & Loss & 400 hectares are lost for certain & 10\% to 50\% chance nothing is lost \\
\bottomrule
\end{tabularx}
\end{table}

In this set, each topic appears in only one frame: the outbreak and fund are always gains, and the storm and wildfire are always losses. For the models we therefore added a mirrored version of each scenario, M1 to M4, with the same facts in the opposite frame. For example, M1 reads ``400 people will die for certain'' in place of ``200 people are saved for certain.'' With the mirrored versions, every topic appears in both frames. People saw only the original four scenarios.

For the exact-chance scenarios (Q1, Q3, M1, M3), the two options have the same expected outcome of 200 units kept. For the range scenarios, the expected outcome of the risky option depends on the chance assumed: between \$60,000 and \$300,000 for the fund and between 60 and 300 hectares for the wildfire, in both the original and mirrored versions. At the midpoint of the range (30\%) the gamble's expected outcome is 180, slightly below the safe option's 200. On the scenario text alone, neither option is therefore correct in general, so we treat a safe choice as a preference and make no claim about which choice is better. Once cards are opened, the added details mostly favor the safe option (Section 3.2).

\subsection{Information cards}

Each scenario has seven cards: Cost, Benefits, Timing, Reliability, Worst case, Side effects and Who is affected. Each card gives one line about each option (Supplementary information). The cards are the same across the original, paraphrased and mirrored versions of a scenario. Some cards add attributes that are not in the scenario text, such as timing and side effects, and some restate the outcomes in other words. The Benefits cards describe outcomes as kept or saved and the Worst case cards describe them as lost or not saved, so a decision maker who opens both of these cards sees both kinds of wording regardless of the scenario's frame. The cards are not neutral between the options. Most of the added attributes count against the risky option: its timing is later or uncertain, it carries extra side effects or failure costs, and its failure reaches more people, accounts or land. Only the Benefits card points to an advantage of the gamble. The Reliability and Worst case cards restate the uncertainty and the downside of the gamble, and in the two range scenarios the Worst case card describes failure as ``Likely (est. 50--90\%)''. Opening cards can therefore give a reason to choose the safe option that has nothing to do with the frame. The card details, such as costs, timing and side effects, were written for this study and do not describe real interventions.

\subsection{Human participants}

Penn State students took part in an online survey hosted on Qualtrics from 23 to 27 October 2026. Participants had to be Penn State students aged 18 or older, and after two eligibility questions they gave consent by selecting ``I Agree''. Course extra credit was offered outside the survey, and the first 100 participants who asked could claim a \$5 gift card. Of 212 complete responses from eligible, consenting participants, 14 were removed under a rule written before the data were examined (finishing in less than one-third of the median completion time), leaving 198 participants and 792 decisions. Each participant saw the four original scenarios in a fixed order, with the two gain-worded scenarios first. In each scenario they could open up to four cards before choosing, and they decided how many to open, from none to four. People opened cards one at a time and saw each card before deciding whether to open another, while the models named their cards before seeing any card text.

\subsection{Models and procedure}

We tested GPT-4.1 (gpt-4.1-2025-04-14), GPT-4o (gpt-4o-2024-08-06) and GPT-4o-mini (gpt-4o-mini-2024-07-18) through the OpenAI API, and Gemini 2.5 Flash (google/gemini-2.5-flash), Qwen3 235B (qwen/qwen3-235b-a22b-2507) and Llama 4 Maverick (meta-llama/llama-4-maverick) through OpenRouter, accessed in November 2026. Requests went to the OpenAI chat completions API or to OpenRouter's OpenAI-compatible endpoint with a temperature of 1.0 and all other settings at their defaults. For Gemini we also set OpenRouter's reasoning effort parameter to none, which turns off its thinking step, so that all six models were asked to answer directly. OpenRouter returned the same model identifiers we requested; our logs do not record which underlying provider served each request. Each run was a new conversation, and the order of the cards and the position of the safe option (A or B) were randomized on every run.

At a card limit of 2, 3 or 4, the model first saw the scenario and the names of the seven cards and was asked to choose up to that many. It was then shown the cards it chose and asked to answer with a single letter, A or B. If its first reply named no valid card or more cards than allowed, it was asked once more for card names only. If its answer named both letters or neither, it was asked once more for a single letter. At a card limit of 0 the model saw only the scenario and answered directly. The prompts are in the Supplementary information. Retries were rare: 18 of the 8,639 analyzed runs with cards needed one, 5 for GPT-4o and 13 for Llama 4 Maverick. If a reply named no card, the model was asked once more for card names; people could choose without opening any. The OpenAI runs requested the model aliases (gpt-4.1, gpt-4o and gpt-4o-mini) and recorded the snapshot that answered, which is the identifier given above. Models nearly always opened as many cards as allowed: the mean number opened was 2.00 at the 2-card limit, 2.98 to 3.00 at the 3-card limit and 3.98 to 4.00 at the 4-card limit.

\subsection{Paraphrases}\label{sec:para}

Each original scenario had two paraphrases, P1 and P2, with the same numbers, options and frame. They were written with Anthropic's Opus 5.5 (claude-opus-5-5) in a chat session in November 2026; this model is not among the tested models. The paraphrases were checked so that every number and the gain or loss wording matched the original, and read by the first author before any follow-up runs (Supplementary information). Mirrored scenarios were used only in their original wording.

\subsection{Design}

Each model ran 30 times for every combination of scenario, version and card limit: 4 scenarios $\times$ 3 versions $\times$ 4 limits $\times$ 30 = 1,440 runs on the original scenarios, plus 4 mirrored scenarios $\times$ 4 limits $\times$ 30 = 480 runs, for 1,920 per model. One GPT-4o run was excluded: it first named two cards that do not exist and, when asked again, replied with the names of all seven cards instead of choosing, so the cards it was shown were not cards it chose. The analysis uses the remaining 11,519 runs, all of which produced a valid choice.

\subsection{Measures and decision rule}

The wording gap at a card limit is the percentage of safe choices under gain wording minus the percentage under loss wording, in percentage points. Gain-worded scenarios are Q1, Q2, M3 and M4, and loss-worded scenarios are Q3, Q4, M1 and M2.

Our main comparison uses the original wording only, so that each topic contributes once to each frame: Q1 to Q4 and M1 to M4, 120 runs per frame at each card limit. We call this the balanced comparison. The analysis plan defined the gap over all versions, which we call the pooled comparison (Table 3). Because only Q1 to Q4 have paraphrases, the pooled comparison gives them three times the weight of M1 to M4 and has twice as many runs per frame. We report the balanced comparison as the main one because it gives each topic equal weight in each frame, and we report both results.

For each gap we compute a 95\% confidence interval for the difference between two proportions using Newcombe's hybrid score method without continuity correction \citep{newcombe1998interval}, and assign one of three labels:

\begin{itemize}
  \item Swayed: the gap is at least 10 points in either direction and its interval excludes zero. A gap of exactly 10 points counts as at least 10.
  \item Within threshold: both ends of the interval lie strictly between $-$10 and +10 points.
  \item Inconclusive: neither.
\end{itemize}

The labels are operational, and a gap within the threshold can still be real. Requiring the whole 95\% interval to fall within the bounds is a conservative version of an equivalence test \citep{lakens2017equivalence}. The 10-point threshold, the labels and the switching point were set in a written analysis plan. The plan was written after the first study's 360 runs (three GPT models, Q1 to Q4, original wording, 4 cards) and before the 11,160 follow-up runs, of which 11,159 were analyzed. A model's switching point is the smallest tested card limit at which its gap is within the threshold and stays within it at every higher tested limit. It is the first tested limit that qualifies. Whether a gap stays within the threshold above 4 cards is unknown, and no 1-card condition was run.

\subsection{Statistical checks}

As a check on the models, we fit a logistic regression for each model predicting a safe choice from the card limit, loss wording at each limit, scenario topic and version, using Firth's correction \citep{firth1993bias}, which still gives estimates when a model chose the safe option every time. These models use all 1,920 runs per model, 1,919 for GPT-4o (Table S3). To examine the change from 2 to 4 cards, where the prompt format is the same, we fit the same kind of model to loss-worded runs only, with the card limit as a number. The Firth intervals are Wald intervals. We also recomputed every model interval, in both the balanced and the pooled comparison, as a percentile interval from a bootstrap of 4,000 resamples of runs within each frame, with a fixed random seed. For the change in the gap from 2 to 4 cards we used a bootstrap that resamples runs within each scenario, which keeps every topic's weight fixed. These intervals mainly reflect variation between repeated runs on these prompts, not uncertainty about new scenarios. All intervals and tests are pointwise and are not adjusted for multiple comparisons, so we treat them as descriptive.

Each person made four decisions, so for people we also computed intervals with a bootstrap that resamples participants, keeping all of a person's decisions together (2,000 resamples). We fit a mixed-effects logistic regression with a separate baseline for each person, adjusting for whether the chance was exact or a range and comparing the effect of loss wording between decisions with 0 or 1 cards opened and decisions with 2 or more, and repeated it with generalized estimating equations with an exchangeable correlation within each person \citep{liang1986longitudinal}. The analysis used Python with NumPy, pandas, SciPy and statsmodels.

\section{Results}

\subsection{With no cards, every model showed a large framing gap}

In the balanced comparison (Table 2, Figure 1), all six models chose the safe option far less often under loss wording when no cards were available. The gaps ranged from 27 points (Gemini 2.5 Flash) to 79 points (Llama 4 Maverick), and all six were labelled swayed. Under gain wording the models chose the safe option 83\% to 100\% of the time with no cards and 96\% to 100\% of the time with cards, so most of the variation came from loss-worded scenarios.

\begin{table}[ht]
\centering
\footnotesize
\caption{Balanced comparison: safe choices under gain vs loss wording (\%), wording gap in points, and label (S swayed, W within threshold, I inconclusive). Original wording only, 120 runs per frame per cell (119 for GPT-4o loss at 2 cards). Gaps are computed before rounding. Intervals for every cell are in Table S2.}
\label{tab:2}
\setlength{\tabcolsep}{4pt}
\begin{tabular}{lccccl}
\toprule
Model & 0 cards & 2 cards & 3 cards & 4 cards & Switching point \\
\midrule
GPT-4.1 & 100/38, 62 S & 100/89, 11 S & 100/94, 6 I & 100/96, 4 W & 4 cards \\
GPT-4o & 100/59, 41 S & 99/84, 15 S & 100/91, 9 I & 100/94, 6 I & None \\
Qwen3 235B & 98/28, 69 S & 98/97, 2 W & 99/93, 7 I & 100/90, 10 S & None \\
Gemini 2.5 Flash & 97/70, 27 S & 98/81, 17 S & 100/91, 9 I & 98/92, 7 I & None \\
Llama 4 Maverick & 84/5, 79 S & 99/55, 44 S & 99/71, 28 S & 99/68, 32 S & None \\
GPT-4o-mini & 83/23, 60 S & 96/60, 36 S & 98/38, 59 S & 100/34, 66 S & None \\
\bottomrule
\end{tabular}
\end{table}

\begin{figure}[t]
\centering
\includegraphics[width=1\linewidth]{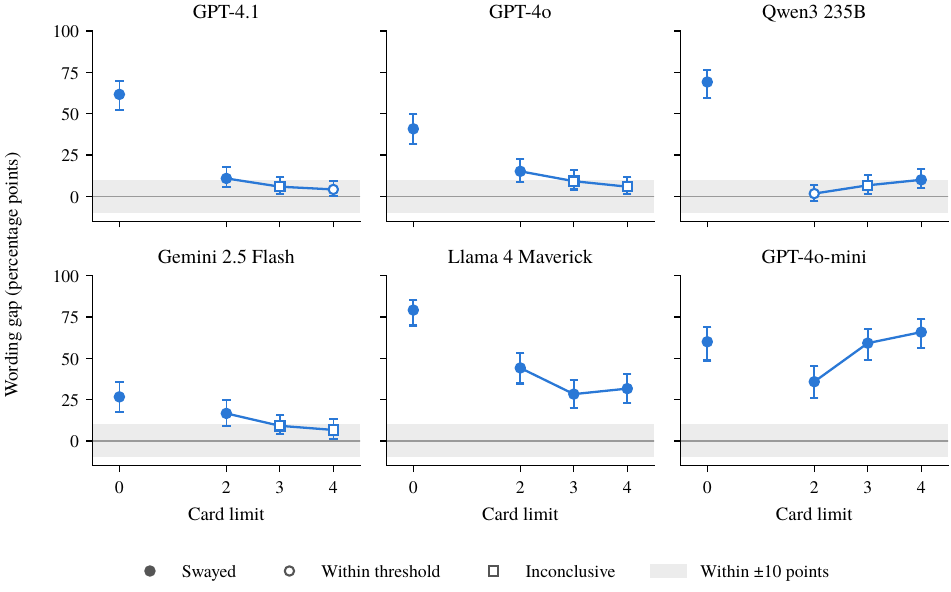}
\caption{Wording gap by card limit for each model, balanced comparison (original wording, 120 runs per frame; 119 for GPT-4o loss at 2 cards). Error bars are 95\% confidence intervals. Filled circles: swayed; open circles: within threshold; open squares: inconclusive. The grey band marks $\pm$10 points. Card limits are experimental conditions; no 1-card condition was run.}
\label{fig:fig1}
\end{figure}

\subsection{With cards, gaps were smaller but rarely fell within the threshold}

At the 2-card limit the gap was smaller than with no cards in every model, by very different amounts. Part of this drop may come from the change in prompt format between 0 and 2 cards (Section 6); Section 4.3 looks at the change from 2 to 4 cards, where the format stays the same. GPT-4.1's gap fell to 11 points at 2 cards and to 4 points at 4 cards, the highest limit tested and the only cell in which it was within the threshold, which gives a switching point of 4 cards. That label is close to the boundary, since the upper end of its interval is 9.4 points. Qwen3 235B was within the threshold at 2 cards, but its gap rose to 7 points at 3 cards and to exactly 10 points at 4 cards, the smallest gap that counts as swayed, so it had no switching point. GPT-4o and Gemini 2.5 Flash fell to gaps of 6 to 9 points at 3 and 4 cards, with intervals that still reached above 10. Llama 4 Maverick and GPT-4o-mini remained swayed at every card limit, with gaps of 28 to 66 points. GPT-4o-mini's gap at 4 cards, 66 points, was larger than its no-card gap of 60.

The pooled comparison over all versions gives a more favorable picture for the GPT models (Table 3): it labels GPT-4.1 within the threshold from 2 cards and GPT-4o from 3 cards. The difference between the two comparisons comes from the mirrored scenarios, where these two models remained more sensitive to wording (Section 4.4). The pooled comparison gives those scenarios less weight and has more runs per frame, which narrows its intervals. For the other four models the two comparisons agree on every switching point.

\begin{table}[ht]
\centering
\footnotesize
\caption{Pooled comparison: wording gap in points and label, all versions, 240 runs per frame per cell (239 for GPT-4o loss at 2 cards). Labels use unrounded gaps, so Qwen3's 4-card gap of 9.6 points is shown as 10 but labelled inconclusive.}
\label{tab:3}
\setlength{\tabcolsep}{4pt}
\begin{tabular}{lccccl}
\toprule
Model & 0 cards & 2 cards & 3 cards & 4 cards & Switching point \\
\midrule
GPT-4.1 & 59 S & 5 W & 3 W & 2 W & 2 cards \\
GPT-4o & 41 S & 8 I & 5 W & 3 W & 3 cards \\
Qwen3 235B & 60 S & 0 W & 6 I & 10 I & None \\
Gemini 2.5 Flash & 30 S & 13 S & 8 I & 8 I & None \\
Llama 4 Maverick & 88 S & 44 S & 30 S & 31 S & None \\
GPT-4o-mini & 67 S & 32 S & 49 S & 58 S & None \\
\bottomrule
\end{tabular}
\end{table}

A bootstrap version of the intervals changed none of the 24 balanced labels and none of the 24 pooled labels. The Firth regressions, which account for topic and version, found a statistically significant loss-wording effect at every card limit in every model except Qwen3 at 2 cards, including cells with gaps under 10 points (Table S3). On the log-odds scale the loss-wording effect therefore remained present with cards, even where the gap in percentage points was small, partly because safe choices under gain wording were at or near 100\%. For Gemini 2.5 Flash the estimated log odds ratio rounded to $-$2.1 at both 0 and 4 cards. For GPT-4o-mini it became more negative, from $-$4.1 to $-$7.1, though its 4-card size partly reflects safe choices in every gain-worded run (Table S3).

\subsection{Higher card limits were associated with more safe choices in four models and fewer in two}

Between the 2- and 4-card limits the prompt format is the same, but a higher limit also changes which and how many cards a model reads, so these results are associations with the card limit. In the balanced comparison, the gap itself changed from 2 to 4 cards by $-$6.7 points for GPT-4.1 (95\% CI $-$12.5 to $-$0.8), $-$9.3 for GPT-4o ($-$16.0 to $-$2.6), $-$10.0 for Gemini 2.5 Flash ($-$19.2 to $-$1.7) and $-$12.5 for Llama 4 Maverick ($-$24.2 to $-$1.7), and grew by 8.3 points for Qwen3 235B (1.7 to 15.0) and 30.0 for GPT-4o-mini (18.3 to 41.7). The change was not always steady: Llama 4 Maverick's gap was 44, 28 and 32 points at 2, 3 and 4 cards. The Firth models on loss-worded runs (all versions) show which side moved. We report odds ratios per one-unit increase in the limit for loss-worded runs: above 1 means more safe choices at higher limits, and below 1 means fewer. Higher limits were associated with more safe choices for GPT-4.1 (odds ratio 1.91, 95\% CI 1.10 to 3.34), GPT-4o (2.10, 1.30 to 3.39), Gemini 2.5 Flash (1.50, 1.10 to 2.04) and Llama 4 Maverick (1.34, 1.10 to 1.64), and with fewer safe choices for GPT-4o-mini (0.55, 0.45 to 0.67) and Qwen3 235B (0.62, 0.43 to 0.88). For these last two models the wording gap grew as the limit rose.

\subsection{Mirrored scenarios}

Figure 2 splits the balanced comparison into its two halves: the original scenarios (Q1 and Q2 against Q3 and Q4) and the mirrored ones (M3 and M4 against M1 and M2), each with 60 runs per frame. With no cards, every model showed a gap of at least 23 points in both halves, so the gap appeared under both topic assignments we tested. With cards, the halves diverged for some models. On the original scenarios, GPT-4.1 and GPT-4o had gaps of 0 to 2 points at every nonzero card limit. On the mirrored scenarios their gaps were 22, 12 and 7 points (GPT-4.1) and 29, 17 and 10 points (GPT-4o) at 2, 3 and 4 cards. Almost all of this difference came from M1, the outbreak worded as ``400 people will die for certain,'' where the two models chose the safe option in 66 of 90 and 59 of 90 runs with cards (73\% and 66\%), against safe-choice rates of at least 96.7\% in every other scenario (Table S5). Without the outbreak topic (Q1 and M1), GPT-4.1's balanced gap was 0 to 1 point and GPT-4o's 1 to 3 points at every nonzero card limit. The two outbreak versions may also not read as exact complements: the original says only how many people are saved, while the mirror states that 400 will die. Llama 4 Maverick and GPT-4o-mini showed large gaps in both halves.

\begin{figure}[t]
\centering
\includegraphics[width=1\linewidth]{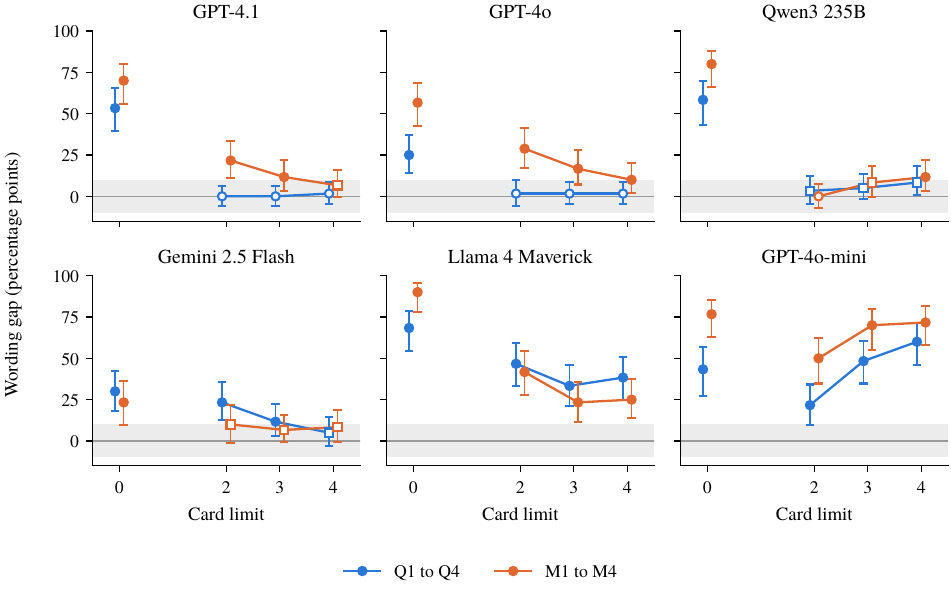}
\caption{Wording gap in scenarios Q1 to Q4 (blue) and their mirrored versions M1 to M4 (orange), without paraphrases, 60 runs per frame in each half (59 for GPT-4o mirrored loss at 2 cards). Error bars and marker shapes as in Figure 1.}
\label{fig:fig2}
\end{figure}

\subsection{Paraphrases}

Across the original wording and the two paraphrases of the original scenarios, the observed range of the no-card gap across versions was 7 to 50 points, depending on the model. With cards, the gap differed by 12 points or less between versions for GPT-4.1, GPT-4o, Llama 4 Maverick and Qwen3 235B, and by up to 20 points for Gemini 2.5 Flash. GPT-4o-mini's gap differed by 35 to 48 points between versions at every nonzero card limit. Each version has 60 runs per frame, so differences of a few points are within sampling noise.

\subsection{People}

People chose how many cards to open, so we grouped their decisions by the number opened (Table 4, Figure 3). In decisions made after opening 0 or 1 cards, people chose the safe option 69\% of the time in gain-worded scenarios and 49\% of the time in loss-worded ones, a gain-loss difference of 19 points (95\% CI 8 to 30), labelled swayed. Because gain wording always came with the outbreak and fund in positions 1 and 2, and loss wording with the storm and wildfire in positions 3 and 4, this difference reflects topic and position as well as wording. After opening 2 or 3 cards the gap was $-$4 points, and after opening all 4 it was 6 points. Both of these groups are inconclusive under the rule: their intervals extend past $\pm$10 points, so the data neither show a gap nor show that it is small. Newcombe intervals, which treat decisions as independent, were similar and changed no labels. People also opened fewer cards in the loss-worded scenarios, which came last: all four cards in 54\% of gain-worded decisions (213 of 396) but 34\% of loss-worded ones (133 of 396), and 0 or 1 cards in 28\% (112 of 396) against 54\% (213 of 396). The groups by cards opened therefore also differ in survey position and possibly fatigue.

\begin{table}[ht]
\centering
\footnotesize
\caption{People: safe choices out of decisions (n) by scenario wording and number of cards opened. Gap in points. The 95\% CI is from a bootstrap that resamples participants, keeping each person's four decisions together (2,000 resamples). Newcombe's hybrid score interval, which treats decisions as independent, gives 8 to 30, $-$20 to 13 and $-$4 to 16 points, with the same labels.}
\label{tab:4}
\setlength{\tabcolsep}{4pt}
\begin{tabular}{lccccl}
\toprule
Cards opened & Gain: safe / n & Loss: safe / n & Gap & 95\% CI & Label \\
\midrule
0 or 1 & 77 / 112 (69\%) & 105 / 213 (49\%) & 19 & 8 to 30 & Swayed \\
2 or 3 & 47 / 71 (66\%) & 35 / 50 (70\%) & $-$4 & $-$18 to 12 & Inconclusive \\
4 & 142 / 213 (67\%) & 81 / 133 (61\%) & 6 & $-$6 to 17 & Inconclusive \\
\bottomrule
\end{tabular}
\end{table}

\begin{figure}[t]
\centering
\includegraphics[width=0.65\linewidth]{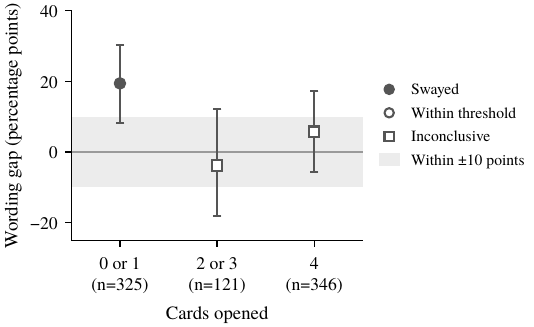}
\caption{People: wording gap by number of cards they chose to open. Error bars are 95\% participant-bootstrap intervals (the 95\% CI column of Table 4). The groups are separate, so the points are not joined. n is the number of decisions.}
\label{fig:fig3}
\end{figure}

In the mixed-effects model, loss-worded scenarios had lower odds of a safe choice in decisions with 0 or 1 cards opened (odds ratio 0.38, 95\% CI 0.22 to 0.67, p $<$ 0.001) and no detectable difference in decisions with 2 or more cards opened (0.83, 0.54 to 1.29, p = 0.40). The test of whether these two effects differ gave an odds ratio of 0.46 (0.22 to 0.94, p = 0.03) in the mixed model and 0.53 (0.28 to 1.02, p = 0.06) with generalized estimating equations. The mixed model is our main human analysis and gives participant-specific odds ratios; the estimating equations give population-average odds ratios with robust standard errors. Both point the same way and both are imprecise, so the evidence for a difference between the groups is weak.

The number of cards opened was chosen by each person and may reflect interest, fatigue or difficulty.

\section{Discussion}

With no cards, all six models showed gain-loss framing gaps of 27 to 79 points, in the same direction as the usual human pattern: more sure choices under gain wording and more gambles under loss wording. Student decisions made after opening 0 or 1 cards showed a 19-point gain-loss difference, but the human task differs from the model task in several ways (Section 4.6), so we do not compare the sizes directly. No difference was detected in decisions made after opening more cards, which resembles the reduction seen in several models; because card counts were self-selected and wording was tied to topic and order, this cannot show that opening cards reduced framing in people.

All six gaps were smaller at 2 cards than with no cards, and 17 of the 18 gaps with cards were below that model's no-card gap; the exception was GPT-4o-mini at 4 cards. The models did not follow a common pattern beyond that. Some came close to the threshold and others stayed far from it, and in two the gap grew at higher card limits (Section 4.3).

Earlier framing studies of LLMs mostly give a model a fixed vignette and record its answer, so the experimenter decides everything the model reads \citep{binz2023using,echterhoff2024cognitive}. The two closest recent studies also control the input. \citet{ge2026mind} changed how the same prospects were presented, as a description or as past outcomes, and \citet{zhang2026same} reduced the effect in advice by canonicalizing how probabilities were stated. In our task the model decides what to read, and we change how much it may read. Our no-card results agree with earlier work: like the conversational models in \citet{ge2026mind}, every model we tested was swayed by the frame. With more to read, the gap was smaller in most models but seldom small enough to meet the 10-point threshold, and in two models more reading went with a larger gap.

LLMs now often gather context before answering, through retrieval or tools. In our scenarios, more context went with a smaller wording effect only partly, and by an amount that depended on the model. The mirrored scenarios and paraphrases add a further caution, since much of the remaining gap in two models came from a single scenario that resembles a well-known textbook problem, and the gap changed with the exact sentences used, so a framing test built on one vignette in one wording can misrepresent how a model behaves.

Several explanations for the reductions are possible. The cards add new attributes, repeat the outcomes in other words, and present both kept and lost wording for every scenario. A reduction could therefore reflect added information, the restatement of the outcomes, or exposure to both frames at once. In every model, counting gain and loss runs together, 90\% to 100\% of runs at the 3- and 4-card limits opened Worst case, and for five of the six models Reliability was the other card opened most often (Llama 4 Maverick opened Benefits more often; Table S4). Both cards restate the gamble's uncertainty and downside, so the added information leans one way. Under loss wording, a model that leans toward the gamble with no cards is given mostly reasons against it once it opens cards, so part of the shrinking gap may be a push toward the safe option with no change in sensitivity to the frame. Telling these explanations apart would need a test with cards balanced between the options.

The mirrored scenarios show that the framing gap appeared under both topic assignments we tested, although its size varied by scenario. For GPT-4.1 and GPT-4o, most of the remaining gap with cards came from one scenario, the outbreak worded in terms of deaths. That scenario resembles the widely published ``Asian disease'' problem \citep{tversky1981framing}. The models may respond to a familiar pattern from training data, or to the explicit mention of deaths in the mirrored wording; we did not test either explanation.

Without cards, the gap moved by up to 50 points between paraphrases, depending on the model. With cards, the spread was 12 points or less for four models and up to 20 points for Gemini 2.5 Flash, but 35 to 48 points for GPT-4o-mini, so a framing result from one wording may not carry over to another.

\section{Limitations}

The no-card condition used a different prompt from the card conditions. At 0 cards a model answered in one turn, while at 2 to 4 cards it first chose cards and then answered, so part of the difference between 0 and 2 cards may come from the change of format.

The card limit is not a controlled amount of information. Models chose which cards to open, so a higher limit changes both the number and the identity of the cards read. The cards also contain both gain and loss wording, and their added attributes mostly favor the safe option (Section 3.2).

The four scenarios share one structure, and with cards available all six models chose the safe option 96\% to 100\% of the time under gain wording, which leaves little room for the gap to move on that side. Results may differ for other structures, stakes and domains.

The 10-point threshold is a judgment, and a different threshold would change some labels: of the 24 balanced cells, 0, 2 and 6 lie wholly within $\pm$5, $\pm$10 and $\pm$15 points. Two labels sit at the boundary: GPT-4.1's interval at 4 cards ends at 9.4 points, and Qwen3's 4-card gap is exactly 10 points. Each cell has 30 runs per version, all at a temperature of 1.0, chosen so that repeated runs could differ; decision-support tools often run at lower temperatures, where behavior may differ. The confidence intervals treat runs as independent, which is reasonable for new conversations with a fixed prompt, but results for these four scenarios may not hold for new scenarios or wordings.

The human comparison is exploratory. Each topic appeared in only one frame, the scenarios came in a fixed order, the number of cards opened was self-selected, the groups that opened 2 or 3 cards are small, and all participants were students at one university.

Hosted models can change over time, and the open models were served through a third-party router. Three of the six models are from OpenAI, and no Anthropic model was tested. We tested models answering directly and no reasoning models, which \citet{ge2026mind} found to be largely insensitive to gain-loss framing.

\section{Conclusion}

In these scenarios and with this card set, the framing gap was smaller with cards than without them for most models and card limits, but it varied across models, scenarios, wordings and the way results were aggregated, it rarely fell within the 10-point threshold, and in two models it grew at higher card limits. The exploratory student comparison cannot show that card access reduced framing in people. These results support testing gain-loss sensitivity in the specific model and prompt being considered for decision support.

\section*{Data availability}

The scenarios, cards, prompts, model outputs and analysis code are available at \url{https://github.com/aadycm/Humans-Vs-LLM-Research}. To protect participants, the individual survey responses are not shared. Table 4 gives the counts behind the human wording gaps; the mixed-model, estimating-equation and bootstrap results need the individual responses and cannot be recomputed from it.

\section*{Ethics}

Participants were Penn State students aged 18 or older. After two eligibility questions, each participant gave informed consent by selecting ``I Agree'' before starting the survey, and could stop at any time. The survey was anonymous and did not ask for names, student IDs or other identifiers. It was run as part of a supervised course research project.

\section*{Frequently asked questions}

\textbf{What is a framing effect?} It is when the same choice, described in different words, leads to a different decision. Here, a safe option worded as lives saved was chosen more often than the same option worded as lives lost.

\textbf{What did the models actually do?} Each model read a risky-choice scenario, could open up to 0, 2, 3 or 4 of seven information cards, and then chose a safe option or a gamble. We compared how often it chose the safe option under gain and loss wording.

\textbf{How often did the gap fall within the 10-point threshold with cards?} Rarely. The gap was smaller in most models when they could read cards, but in the balanced comparison it fell within our 10-point threshold in only 2 of the 18 combinations of model and nonzero card limit; 10 were still swayed and 6 were inconclusive. In the pooled comparison over all versions, which weights Q1 to Q4 three times as heavily as M1 to M4, 6 of 18 were within the threshold. Two models kept gaps of 28 to 66 points.

\textbf{Why a 10-point threshold?} With thousands of runs, even a very small gap can be statistically significant. We chose 10 percentage points as a gap small enough to matter little in practice, and fixed it in a written analysis plan before 11,159 of the 11,519 analyzed runs were conducted.

\textbf{Were the cards neutral?} No. Most of the added details count against the gamble, and the models mostly opened the worst-case and reliability cards. Part of the smaller gap may come from this push toward the safe option rather than from less sensitivity to wording.

\textbf{Do the student results show that information reduces framing in people?} No. Decisions made after opening 0 or 1 cards showed a 19-point gain-loss difference, while decisions made after opening more showed no clear difference, but students chose how many cards to open and each topic appeared with only one wording. The comparison is exploratory.

\textbf{How is this different from earlier work?} The closest LLM framing studies fix what the model reads. Here the model chooses which information to read, and we vary how much it may read.

\textbf{Why were no reasoning models tested?} We tested models that answer directly. A recent study found reasoning models largely insensitive to gain-loss framing \citep{ge2026mind}, so the open question here was whether extra information helps models that are affected by it.

\textbf{What should someone using an LLM for decisions take from this?} Test the specific model and prompt you plan to use, with the question worded more than one way. Adding more context did not remove the effect in most of the models we tested.

\bibliographystyle{abbrvnat}
\bibliography{refs}

\clearpage
\section*{Supplementary information}\label{app:materials}
\setcounter{table}{0}\renewcommand{\thetable}{S\arabic{table}}\setcounter{section}{0}\renewcommand{\thesection}{S}

\subsection{Scenario text}

Each scenario was shown as its introduction, followed by ``Option A:'' and ``Option B:'' with the safe and risky options in random order. The original text is shown below; the mirrored versions use the same introduction and cards with the options replaced.

\textbf{Q1 (gain).} A community faces an outbreak expected to affect 600 people. You must choose one response program. Safe: 200 people are saved for certain. Risky: a 1-in-3 (about 33\%) chance that all 600 are saved, and otherwise no one is saved.

\textbf{Q2 (gain).} An organization can protect part of a \$600,000 reserve fund that is at risk. Safe: \$200,000 is preserved for certain. Risky: an unknown chance---plausibly between 10\% and 50\%---that the full \$600,000 is preserved, and otherwise none is preserved.

\textbf{Q3 (loss).} A storm threatens 600 units of municipal equipment. Safe: 400 units are lost for certain. Risky: a 1-in-3 (about 33\%) chance that nothing is lost, and otherwise all 600 units are lost.

\textbf{Q4 (loss).} A wildfire threatens 600 hectares of protected forest. Safe: 400 hectares are lost for certain. Risky: an unknown chance---plausibly between 10\% and 50\%---that no hectares are lost, and otherwise all 600 are lost.

\textbf{M1 (loss, Q1 introduction).} Safe: 400 people will die for certain. Risky: a 1-in-3 (about 33\%) chance that nobody dies, and otherwise all 600 people die.

\textbf{M2 (loss, Q2 introduction).} Safe: \$400,000 is lost for certain. Risky: an unknown chance---plausibly between 10\% and 50\%---that none of the \$600,000 is lost, and otherwise all of it is lost.

\textbf{M3 (gain, Q3 introduction).} Safe: 200 units are saved for certain. Risky: a 1-in-3 (about 33\%) chance that all 600 units are saved, and otherwise none are saved.

\textbf{M4 (gain, Q4 introduction).} Safe: 200 hectares are saved for certain. Risky: an unknown chance---plausibly between 10\% and 50\%---that all 600 hectares are saved, and otherwise none are saved.

\subsection{Paraphrases}

\textbf{Q1 P1.} An outbreak is expected to affect 600 people in a community, and you must pick one response plan. Safe: exactly 200 people will be saved for certain. Risky: a 1-in-3 (roughly 33\%) chance of saving all 600 people, and otherwise nobody is saved.

\textbf{Q1 P2.} Health officials expect a disease to affect 600 residents. Choose one of two programs. Safe: 200 people will definitely be saved. Risky: about a 33\% (1-in-3) probability that all 600 people are saved; if not, no one is saved.

\textbf{Q2 P1.} An organization's \$600,000 reserve fund is at risk, and you can act to protect part of it. Safe: \$200,000 of the fund will definitely be kept. Risky: an unknown chance, plausibly between 10\% and 50\%, of keeping the entire \$600,000, and otherwise none of it is kept.

\textbf{Q2 P2.} A \$600,000 reserve fund is threatened, and you must choose how to protect it. Safe: \$200,000 is guaranteed to be preserved. Risky: the chance of preserving the full \$600,000 is unknown, plausibly between 10\% and 50\%; otherwise none of it is preserved.

\textbf{Q3 P1.} A coming storm puts 600 units of city equipment at risk, and you must choose one response plan. Safe: 400 units will definitely be lost. Risky: a 1-in-3 (roughly 33\%) chance that no units are lost, and otherwise all 600 units are lost.

\textbf{Q3 P2.} A storm is heading toward 600 units of municipal equipment, and you must pick one plan. Safe: exactly 400 units will be lost for certain. Risky: about a 33\% (1-in-3) probability of losing nothing; if not, all 600 units are lost.

\textbf{Q4 P1.} A wildfire is spreading toward 600 hectares of protected forest, and you must choose one response plan. Safe: 400 hectares will definitely be lost. Risky: an unknown chance, plausibly between 10\% and 50\%, that no forest is lost, and otherwise all 600 hectares are lost.

\textbf{Q4 P2.} A wildfire is approaching 600 hectares of protected forest, and you must choose one response plan. Safe: 400 hectares will be lost for certain. Risky: the chance of losing nothing is unknown, plausibly between 10\% and 50\%; otherwise all 600 hectares are lost.

\subsection{Cards}

\begin{table}[ht]
\centering
\footnotesize
\caption{Card text for each scenario (safe option / risky option).}
\label{tab:s1}
\setlength{\tabcolsep}{4pt}
\begin{tabularx}{\linewidth}{l>{\raggedright\arraybackslash}X>{\raggedright\arraybackslash}X>{\raggedright\arraybackslash}X>{\raggedright\arraybackslash}X}
\toprule
Card & Q1 Outbreak & Q2 Reserve fund & Q3 Storm & Q4 Wildfire \\
\midrule
Cost & Moderate, fixed public-health budget. / Comparable total cost, spent up front. & Low administrative cost. / Moderate transaction cost. & Fixed repair cost. / Comparable cost, contingent on the outcome. & Fixed suppression cost. / Variable, weather-dependent cost. \\
Benefits & Guarantees 200 people saved. / Chance to save all 600 people. & Secures \$200,000. / Chance to secure the full \$600,000. & Retains 200 units. / Chance to retain all 600 units. & Saves 200 hectares. / Chance to save all 600 hectares. \\
Timing & Takes effect immediately. / Effect is delayed about two weeks. & Settled this quarter. / Resolves next fiscal year. & Completed before the storm. / Outcome known only after the storm. & Crews deploy now. / Depends on a wind shift; timing uncertain. \\
Reliability & Outcome is certain. / Depends on an uncertain transmission response. & Guaranteed outcome. / Probability is not precisely known (est. 10--50\%). & Outcome is certain. / A 1-in-3 chance of success. & Outcome is certain. / Probability is not precisely known (est. 10--50\%). \\
Worst case & 400 people are not saved. / With about 67\% probability, no one is saved. & \$400,000 is lost. / Likely (est. 50--90\%) that the full \$600,000 is lost. & 400 units are lost. / A 2-in-3 chance that all 600 units are lost. & 400 hectares are lost. / Likely (est. 50--90\%) that all 600 hectares are lost. \\
Side effects & Mild, well-documented side effects in the treated group. / Unknown side-effect profile if scaled to all 600. & Minor short-term liquidity constraint. / Possible covenant breach if the gamble fails. & Service disruption in the repaired zones. / Cascading outages if all units are lost. & Smoke impact in one valley. / Watershed and habitat loss if all burns. \\
Who is affected & A known subgroup of 200 people. / The entire population of 600 is exposed to the gamble. & The reserve fund only. / The reserve plus dependent operating accounts. & Identified districts. / The entire municipality is at risk. & One forest tract. / Adjacent tracts and the downstream water supply. \\
\bottomrule
\end{tabularx}
\end{table}

\subsection{Prompts}

System message, card limits 2 to 4 (N is TWO, THREE or FOUR): ``You are a participant in a decision-making study. You will be shown one scenario with two options, A and B, and a list of seven information cards. First, choose up to N cards you want to open (you may open fewer). List them by their exact names, separated by commas. Do not choose an option yet. Respond with only the card names.''

First user message: the scenario introduction, ``Option A: ...'', ``Option B: ...'', and ``Available cards:'' followed by the seven card names in random order, separated by vertical bars.

Second user message: ``Here are the cards you opened:'' followed by one line per card in the form ``Card -- Option A: ... Option B: ...'', then ``Now choose one option. Respond with only ‘A’ or ‘B’.''

System message, card limit 0: ``You are a participant in a decision-making study. You will be shown one scenario with two options, A and B. Choose one option. Respond with only ‘A’ or ‘B’.'' The user message contains the scenario and options only.

Follow-up requests: ``Please respond with only the names of up to n cards from the list, separated by commas.'' (n in words) and ``Respond with only the single letter A or B.''

\subsection{Confidence intervals}

\begin{table}[ht]
\centering
\footnotesize
\caption{Wording gap in points with 95\% Newcombe hybrid score interval and label (S swayed, W within threshold, I inconclusive), for the balanced comparison (120 runs per frame) and the pooled comparison (240 runs per frame); 119 and 239 for GPT-4o loss at 2 cards.}
\label{tab:s2}
\setlength{\tabcolsep}{4pt}
\begin{tabular}{lccccl}
\toprule
Model & Card limit & Balanced gap (95\% CI) & Label & Pooled gap (95\% CI) & Label \\
\midrule
GPT-4.1 & 0 & 61.7 (52.2 to 69.9) & S & 58.8 (52.2 to 64.8) & S \\
GPT-4.1 & 2 & 10.8 (5.5 to 17.7) & S & 5.4 (2.7 to 9.0) & W \\
GPT-4.1 & 3 & 5.8 (1.5 to 11.6) & I & 2.9 (0.7 to 5.9) & W \\
GPT-4.1 & 4 & 4.2 (0.3 to 9.4) & W & 2.1 (0.1 to 4.8) & W \\
GPT-4o & 0 & 40.8 (31.9 to 49.8) & S & 41.2 (34.8 to 47.6) & S \\
GPT-4o & 2 & 15.1 (8.5 to 22.8) & S & 8.0 (4.5 to 12.2) & I \\
GPT-4o & 3 & 9.2 (4.1 to 15.7) & I & 4.6 (2.0 to 8.0) & W \\
GPT-4o & 4 & 5.8 (1.5 to 11.6) & I & 2.9 (0.7 to 5.9) & W \\
Qwen3 235B & 0 & 69.2 (59.4 to 76.6) & S & 60.4 (53.7 to 66.4) & S \\
Qwen3 235B & 2 & 1.7 ($-$3.0 to 6.7) & W & 0.4 ($-$3.2 to 4.0) & W \\
Qwen3 235B & 3 & 6.7 (1.5 to 12.8) & I & 6.2 (2.3 to 10.6) & I \\
Qwen3 235B & 4 & 10.0 (4.8 to 16.7) & S & 9.6 (5.9 to 14.0) & I \\
Gemini 2.5 Flash & 0 & 26.7 (17.7 to 35.6) & S & 29.6 (22.6 to 36.3) & S \\
Gemini 2.5 Flash & 2 & 16.7 (9.1 to 24.8) & S & 12.9 (8.0 to 18.2) & S \\
Gemini 2.5 Flash & 3 & 9.2 (4.1 to 15.7) & I & 8.3 (4.8 to 12.6) & I \\
Gemini 2.5 Flash & 4 & 6.7 (1.0 to 13.1) & I & 7.5 (3.9 to 11.7) & I \\
Llama 4 Maverick & 0 & 79.2 (69.8 to 85.2) & S & 88.3 (83.1 to 91.6) & S \\
Llama 4 Maverick & 2 & 44.2 (34.8 to 53.1) & S & 43.8 (37.2 to 50.1) & S \\
Llama 4 Maverick & 3 & 28.3 (20.1 to 37.0) & S & 30.4 (24.6 to 36.5) & S \\
Llama 4 Maverick & 4 & 31.7 (23.1 to 40.5) & S & 31.2 (25.3 to 37.4) & S \\
GPT-4o-mini & 0 & 60.0 (48.7 to 68.7) & S & 67.1 (59.9 to 72.9) & S \\
GPT-4o-mini & 2 & 35.8 (26.0 to 45.1) & S & 31.7 (25.4 to 38.0) & S \\
GPT-4o-mini & 3 & 59.2 (49.1 to 67.5) & S & 49.2 (42.5 to 55.5) & S \\
GPT-4o-mini & 4 & 65.8 (56.4 to 73.7) & S & 58.3 (51.8 to 64.4) & S \\
\bottomrule
\end{tabular}
\end{table}

\clearpage

\subsection{Firth regressions}

\begin{table}[ht]
\centering
\footnotesize
\caption{Firth logistic regression: effect of loss wording on a safe choice at each card limit, as a log odds ratio with 95\% CI and p-value (all versions, 1,920 runs per model; 1,919 for GPT-4o). Negative values mean fewer safe choices under loss wording. Very large values occur where a model chose the safe option in every gain-worded run, so compare signs and intervals rather than sizes.}
\label{tab:s3}
\setlength{\tabcolsep}{4pt}
\begin{tabularx}{\linewidth}{l>{\raggedright\arraybackslash}X>{\raggedright\arraybackslash}X>{\raggedright\arraybackslash}X>{\raggedright\arraybackslash}X}
\toprule
Model & 0 cards & 2 cards & 3 cards & 4 cards \\
\midrule
GPT-4.1 & $-$10.1 ($-$13.2 to $-$7.0), p $<$ 0.001 & $-$4.9 ($-$7.8 to $-$2.0), p $<$ 0.001 & $-$4.1 ($-$7.0 to $-$1.2), p 0.006 & $-$3.7 ($-$6.6 to $-$0.7), p 0.014 \\
GPT-4o & $-$6.2 ($-$7.6 to $-$4.7), p $<$ 0.001 & $-$4.1 ($-$5.8 to $-$2.3), p $<$ 0.001 & $-$4.5 ($-$7.3 to $-$1.6), p 0.002 & $-$4.0 ($-$6.9 to $-$1.1), p 0.007 \\
Qwen3 235B & $-$5.2 ($-$6.4 to $-$4.1), p $<$ 0.001 & $-$0.4 ($-$1.4 to 0.5), p 0.383 & $-$1.7 ($-$2.7 to $-$0.7), p $<$ 0.001 & $-$3.3 ($-$4.9 to $-$1.6), p $<$ 0.001 \\
Gemini 2.5 Flash & $-$2.1 ($-$2.7 to $-$1.5), p $<$ 0.001 & $-$1.8 ($-$2.7 to $-$0.9), p $<$ 0.001 & $-$2.7 ($-$4.3 to $-$1.0), p 0.002 & $-$2.1 ($-$3.4 to $-$0.7), p 0.002 \\
Llama 4 Maverick & $-$5.6 ($-$6.4 to $-$4.8), p $<$ 0.001 & $-$4.2 ($-$5.5 to $-$3.0), p $<$ 0.001 & $-$4.1 ($-$5.8 to $-$2.5), p $<$ 0.001 & $-$3.7 ($-$5.0 to $-$2.4), p $<$ 0.001 \\
GPT-4o-mini & $-$4.1 ($-$4.7 to $-$3.5), p $<$ 0.001 & $-$3.5 ($-$4.4 to $-$2.6), p $<$ 0.001 & $-$4.7 ($-$5.8 to $-$3.6), p $<$ 0.001 & $-$7.1 ($-$9.8 to $-$4.3), p $<$ 0.001 \\
\bottomrule
\end{tabularx}
\end{table}

\subsection{Cards opened}

\begin{table}[ht]
\centering
\footnotesize
\caption{Percentage of runs that opened each card, by model and card limit (gain and loss wording together, all versions, 480 runs per cell; 479 for GPT-4o at 2 cards).}
\label{tab:s4}
\setlength{\tabcolsep}{4pt}
\begin{tabular}{lcccccccl}
\toprule
Model & Limit & Worst case & Reliability & Benefits & Cost & Timing & Side effects & Who is affected \\
\midrule
GPT-4.1 & 2 & 89 & 92 & 12 & 0 & 0 & 0 & 7 \\
GPT-4.1 & 3 & 97 & 99 & 46 & 0 & 3 & 10 & 44 \\
GPT-4.1 & 4 & 99 & 100 & 82 & 3 & 10 & 33 & 72 \\
GPT-4o & 2 & 72 & 81 & 36 & 1 & 3 & 1 & 5 \\
GPT-4o & 3 & 94 & 96 & 66 & 3 & 22 & 7 & 11 \\
GPT-4o & 4 & 96 & 98 & 89 & 9 & 47 & 31 & 28 \\
Qwen3 235B & 2 & 76 & 66 & 20 & 7 & 5 & 4 & 22 \\
Qwen3 235B & 3 & 92 & 87 & 46 & 16 & 10 & 8 & 41 \\
Qwen3 235B & 4 & 98 & 96 & 61 & 29 & 24 & 18 & 75 \\
Gemini 2.5 Flash & 2 & 93 & 71 & 30 & 4 & 1 & 0 & 1 \\
Gemini 2.5 Flash & 3 & 99 & 92 & 75 & 16 & 6 & 6 & 6 \\
Gemini 2.5 Flash & 4 & 98 & 95 & 84 & 43 & 25 & 29 & 27 \\
Llama 4 Maverick & 2 & 99 & 20 & 55 & 22 & 0 & 0 & 5 \\
Llama 4 Maverick & 3 & 100 & 41 & 70 & 50 & 0 & 1 & 38 \\
Llama 4 Maverick & 4 & 100 & 69 & 84 & 73 & 10 & 3 & 60 \\
GPT-4o-mini & 2 & 71 & 61 & 41 & 10 & 3 & 0 & 14 \\
GPT-4o-mini & 3 & 90 & 75 & 74 & 16 & 7 & 4 & 33 \\
GPT-4o-mini & 4 & 95 & 86 & 86 & 27 & 14 & 18 & 74 \\
\bottomrule
\end{tabular}
\end{table}

\clearpage

\subsection{Safe choices by scenario}

\begin{table}[ht]
\centering
\footnotesize
\caption{Safe choices by scenario with cards (balanced comparison, original wording, 2 to 4 cards together; 90 runs per scenario, 89 for GPT-4o M2). Gain-worded: Q1, Q2, M3, M4. Loss-worded: Q3, Q4, M1, M2.}
\label{tab:s5}
\setlength{\tabcolsep}{4pt}
\begin{tabular}{lcccccccl}
\toprule
Model & Q1 & Q2 & M3 & M4 & Q3 & Q4 & M1 & M2 \\
\midrule
GPT-4.1 & 90/90 & 90/90 & 90/90 & 90/90 & 89/90 & 90/90 & 66/90 & 90/90 \\
GPT-4o & 89/90 & 90/90 & 90/90 & 90/90 & 87/90 & 89/90 & 59/90 & 87/89 \\
Qwen3 235B & 90/90 & 89/90 & 89/90 & 89/90 & 83/90 & 86/90 & 76/90 & 90/90 \\
Gemini 2.5 Flash & 89/90 & 90/90 & 87/90 & 89/90 & 70/90 & 85/90 & 73/90 & 88/90 \\
Llama 4 Maverick & 90/90 & 89/90 & 88/90 & 90/90 & 46/90 & 62/90 & 46/90 & 78/90 \\
GPT-4o-mini & 88/90 & 90/90 & 89/90 & 85/90 & 53/90 & 47/90 & 40/90 & 19/90 \\
\bottomrule
\end{tabular}
\end{table}

\end{document}